\documentclass[runningheads]{llncs}

\usepackage{booktabs} 
\usepackage{graphicx}
\usepackage{colortbl}
\usepackage{subfig}
\usepackage{url}
\usepackage{algorithm,algorithmic}
\usepackage{multicol}
\usepackage{lscape}

\begin{document}

\title{MOEA/D with Adaptative Number of Weight Vectors}
\author{Yuri Lavinas\inst{1} \and
Abe Mitsu Teru\inst{2} \and
Yuta Kobayashi\inst{3} \and
Claus Aranha\inst{4}}
\authorrunning{Lavinas, Y. et al.}
%
\institute{University of Tsukuba, Japan \email{lavinas.yuri.xp@alumni.tsukuba.ac.jp} \and
	University of Tsukuba, Japan \email{abe.mitsuteru.xw@alumni.tsukuba.ac.jp} \and
	University of Tsukuba, Japan \email{kobayashi.yuta.xu@alumni.tsukuba.ac.jp} \and
	University of Tsukuba, Japan \email{caranha@cs.tsukuba.ac.jp}\\
}

\maketitle

\begin{abstract}
The Multi-Objective Evolutionary Algorithm based on Decomposition (MOEA/D) is a popular algorithm for solving Multi-Objective Problems (MOPs). The main component of MOEA/D is to decompose a MOP into easier sub-problems using a set of weight vectors. The choice of the number of weight vectors significantly impacts the performance of MOEA/D. However, the right choice for this number varies, given different MOPs and search stages. 
Here we adaptively change the number of vectors by removing unnecessary vectors and adding new ones in empty areas of the objective space. 
Our MOEA/D variant uses the Consolidation Ratio to decide when to change the number of vectors, and then it decides where to add or remove these weighted vectors. 
We investigate the effects of this adaptive MOEA/D against MOEA/D with a poorly chosen set of vectors, a MOEA/D with fine-tuned vectors and MOEA/D-AWA on the DTLZ and ZDT benchmark functions. We analyse the algorithms in terms of hypervolume, IGD and entropy performance.
Our results show that the proposed method is equivalent to MOEA/D with fine-tuned vectors and superior to MOEA/D with poorly defined vectors. Thus, our adaptive mechanism mitigates problems related to the choice of the number of weight vectors in MOEA/D, increasing the final performance of MOEA/D by filling empty areas of the objective space while avoiding premature stagnation of the search progress.

\end{abstract}



\section{Introduction}


Multi-objective Optimisation Problems (MOPs) involve multiple objectives with trade-off relationships making it hard to find a single solution that provides a good balance. Thus, to satisfy all objectives, there is a need to find a set of trade-off solutions, called Pareto Front (PF) solutions. A critical characteristic of this solutions is to cover all regions of the optimal and continual PF, without any sparsity regions~\footnote{In the case of discontinuous PF case, there is no need to cover the discontinuity area.}. These empty areas of the PF indicate that different possible trade-offs are still to be found.

One of the most common algorithms for finding good sets of solutions for MOPs is the Multi-Objective Evolutionary Algorithm Based on Decomposition (MOEA/D)~\cite{moead_original}. The most crucial feature of MOEA/D is that this algorithm decomposes the MOP into several single-objective problems. This decomposition of the MOP depends on a set of \emph{weight vectors}, where each vector corresponds to a different region of the PF. The choice of weight vectors is essential on MOEA/D and the appropriate value generally is not known. Also, these weight vectors are closely related to the population size, influencing its dynamic during the search progress, depending on the difficulty of the problem, the presence of multiple local optima, the shape of the PF and other features~\cite{vcrepinvsek2013exploration,glasmachers2014start}. Thus, using a low number of vectors may lead to search stagnation, while a high number may waste computational resources.

A growing body of literature recognises the need to define the appropriate set of weight vectors in MOEA/D~\cite{moead_awa,AdaW,moead_uraw,fv_moead,survey_WVA}. One major issue in these works is that they focus on adjusting the position of weight vectors in terms of the objective space, paying little attention to defining the number of weight vectors. In summary, much uncertainty still exists about the relationship between redefining the weight vectors adaptively in MOEA/D and the coverage of the empty spaces of the PF region while avoiding premature convergence.

Here, we focus on automatically adapting the number of weight vectors in MOEA/D, adding or deleting vectors based on the progress of the search. Our proposed adaptation method has two main components: (1) to identify the timing to add and remove weight vectors, and (2) to decide which vectors to add or remove. To identify the timing to change vectors, we use the \textit{Consolidation Ratio}, which was initially proposed as an online stopping criteria~\cite{cr}. To decide which vectors to add or remove, we use two strategies, one based on uniform sampled values and the other based on the adaptive weight adjustment (AWA)~\cite{moead_awa}. Moreover, our method dependency on the initial set of weight vectors is small and given its adaptive nature requiring little fine-tuning of the number of vectors.

The proposed method was tested on the DTLZ and ZDT benchmark and compared with (1) MOEA/D with different weight vectors settings and (2) MOEA/D-AWA~\cite{moead_awa}, a method that adjusts the positions of the weight vectors during the search. This study provides new insights into the relationship between the choice of the number of weight vectors in MOEA/D and the increments of performance of MOEA/D by filling empty areas of the objective space while avoiding premature stagnation of the search progress.



\section{MOEA/D and Weight Vectors}

The MOEA/D algorithm is characterized by the decomposition of the MOP into many sub-problems. A sub-problem is characterized by a weight vector $\lambda$ highly influencing the performance of MOEA/D. When adding new weight vectors, it is necessary to decide where a new set of vectors should be positioned. Several works study the problem of changing the positions of weight vectors in MOEA/D. For more detailed information, refer to this Survey of Ma et al.~\cite{survey_WVA}. One of the most popular methods for guiding the weight adaptation strategy is the Adaptive Weight Adjustment (AWA)~\cite{moead_awa,AdaW,moead_uraw,fv_moead}. A major advantage of AWA is that it changes the position of the vectors according to the feature of the MOPs and that is why we  use AWA as one of the base mechanisms in this work.



The AWA method keeps an external archive and re-position the vectors to the sparsest regions of this archive. This positioning is based on the Sparsity Level (SL) for each solution in the archive:
\begin{equation}
    SL(ind_{j},pop) = \prod_{i=1}^{m}L_{2}^{NN_{i}^{j}}
    \label{eq::vicinty distancs}
\end{equation}
where $L_{2}^{NN_{i}^{j}}$ is the Euclidean distance between the $j$-th solution and its $i$-th nearest neighbour. The, a new vector, $\bf{\lambda^{sp}}$, is generated using the individual with the highest SL as based for calculation. 

\begin{equation}
    \bf{\lambda^{sp}} = \left( \frac{\frac{1}{f_{1}^{sp}-z_{1}^{*}+\epsilon}}{\sum_{k=1}^{m}\frac{1}{f_{k}^{sp}-z_{k}^{*}+\epsilon}},..., \frac{\frac{1}{f_{m}^{sp}-z_{m}^{*}+\epsilon}}{\sum_{k=1}^{m}\frac{1}{f_{k}^{sp}-z_{k}^{*}+\epsilon}} \right)
\label{eq::sparce vector}
\end{equation}

Where $f_{k}^{sp}$ is the objective function value of the individual with the biggest SL, and $\epsilon$ is a small number. This paper uses a modified version that considers the Unbounded External Archive (UEA) instead of the traditional external archive. For more information see Algorithm~\ref{algo::addVector}.


The proposed method uses the stagnation state of the search as criteria to decide when to add or remove weight vectors. Several studies study stagnation metrics in MOEAs as stopping criteria~\cite{cr,ocd}. Here, we choose the CR indicator~\cite{cr}, since it requires no problem-dependent parameters. 

The CR indicator uses the non-dominated archive $A_{i}$ of the population at generation $i$, the non-dominated archive at generation $i-\Delta$, and the set $S$ of solutions from $A_{i-\Delta}$ that are not dominated by $A_{i}$, and is calculated as
\begin{eqnarray}
    CR&=&\frac{|S|}{|A_{i}|},\\  
    S&=&\{a_{i-\Delta}:a_{i-\Delta}\not \prec a_{i}\},\\
    & & a_{i}\in A_{i}, a_{i-\Delta} \in A_{i-\Delta}.
    \label{eq::consolidation ratio}
\end{eqnarray}
Using this indicator, we can compute the utility function $U_{i}$, and the average generation utility $U_{i}^{*}$ (equations~\ref{eq::utility} and~\ref{eq::average generation utility}). The decision to add or remove weights is made when $U_{i}^{*}$ exceeds a threshold $U_{thresh}$, that depends on the $U_{init}$ value.

\begin{equation}
    U_{i} = \frac{CR_{i}-CR_{(i-\Delta)}}{\Delta}
    \label{eq::utility}
\end{equation}

\begin{equation}
    U_{i}^{*} = \frac{U_{i}+U_{(i-\Delta)}}{2}
    \label{eq::average generation utility}
\end{equation}

\begin{equation}
    U_{init} = \frac{CR|_{\gg 0.5}}{G}
    \label{eq::utility init}
\end{equation}

\begin{equation}
    U_{thresh} = \frac{U_{init}}{F}
    \label{eq::utility threshold}
\end{equation}

F and $\Delta$ values were chosen based on the original CR paper~\cite{cr}. $U_{init}$ is calculated once when the CR exceeds $0.5$ for the first time, at generation $G$.
\section{MOEA/D with Adaptive Weight Vectors}

We propose a method to enhance MOEA/D by automatically adding or removing weight vectors as the search progresses, named MOEA/D-AV (\textbf{MOEA/D} with \textbf{A}daptive weight \textbf{V}ectors). The outline of this method is described in Algorithm~\ref{algo::Proposed method}. The code for the method and experiments in this study is available at a GitHub repository~\footnote{\url{https://github.com/YUYUTA/MOEADpy}}.

\begin{algorithm}
\caption{MOEA/D-AV}
\label{algo::Proposed method}
\begin{algorithmic}[1]
\renewcommand{\algorithmicrequire}{\textbf{Input:}}
\renewcommand{\algorithmicensure}{\textbf{Output:}}
\REQUIRE number of adjustment weights $\bf{ratio}$,
  initial weight vectors $\bf{W}$, MOEA/D variables (set of solutions, neighborhood solution matrix,...)
\ENSURE  Unbounded External Archive $\bf{UEA}$

\STATE $\bf{UEA}$ $\leftarrow$  $\emptyset$
\STATE Initialize and evaluate population $\bf{X^{(0)}}$
\STATE Update $\bf{UEA}$
\STATE set $CR_{Gen}$ $\leftarrow$ NULL, $U_{thresh}$ $\leftarrow$ NULL
    \WHILE{termination criterion is not meet}
        \STATE Generate new population $\bf{X'^{(Gen)}}$ and evaluate this new population
        \STATE Update $\bf{UEA}$ \& Select next population
        \IF {$CR_{Gen}$ is NULL}
            \STATE Calculate $CR_{Gen}$.
        \ENDIF
        \IF{$CR_{Gen} > 0.5$}
            \IF{$U_{thresh}$ is NULL}
                \STATE Calculate $U_{thresh}$
            \ENDIF
            \IF{$U_{Gen}^{*} > U_{thresh}$}
                \STATE $p = \frac{n\_fe}{n\_eval}$
                \STATE $\bf{nav} = \bf{ratio}*size(\bf{W})$
                \IF{$p>random$}
                    \STATE Add vectors using the Unbounded version of AWA (Algorithm~\ref{algo::addVector})
                \ELSE
                    \STATE Add vectors using the Uniform selection method (Algorithm~\ref{algo::addVector2})
                \ENDIF
            \ENDIF
        \ELSE
            \STATE Delete vectors (Algorithm~\ref{algo::delete Vector})
        \ENDIF
    \ENDWHILE
\end{algorithmic} 
\end{algorithm}

Now we explain the most relevant details of MOEA/D-AV. At every generation, this algorithm calculates the CR value. When the value of CR is larger than $0.5$, the method moves on to calculate the threshold value $U_{thresh}$. Then at every generation, MOEA/D-AV calculates the average generation utility (equation~\ref{eq::average generation utility}). If it exceeds the value of $U_{thresh}$, then the algorithm adds new weight vectors. Otherwise, it deletes weight vectors. The number of vectors added or removed at each update is decided by the fraction of the total number of weight vectors.

\begin{algorithm}
\caption{Adding Vectors - method 1 (adds vectors based on the sparsity of the UEA)}
\label{algo::addVector}
\begin{algorithmic}[1]
\renewcommand{\algorithmicrequire}{\textbf{Input:}}
\renewcommand{\algorithmicensure}{\textbf{Output:}}
\REQUIRE Unbounded External Archive $\bf{UEA}$, current population {\bf pop}, number of to add vectors {\bf nav}
\ENSURE  Updated population $\bf{pop}$
\STATE set count = 0
\STATE calculate SL of individual in $\bf{UEA}$ using Eq.~\ref{eq::vicinty distancs}
\WHILE{$count < \bf{nav}$}
\STATE set $\bf{ind^{sp}}$ =($\bf{x^{sp}},\bf{FV^{sp}}$) which has largest SL
\STATE generate new vector $\bf{\lambda^{sp}}$ using Eq.~\ref{eq::sparce vector}
\STATE add ($\bf{ind^{sp}}$,$\bf{\lambda^{sp}}$) to $\bf{pop}$
\STATE count = count+1
\ENDWHILE
\RETURN $\bf{pop}$
\end{algorithmic} 
\end{algorithm}

\begin{algorithm}[t]
\caption{Adding Vectors - method 2 (adds vectors based on values sampled from a uniform distribution)}
\label{algo::addVector2}
\begin{algorithmic}[1]
\renewcommand{\algorithmicrequire}{\textbf{Input:}}
\renewcommand{\algorithmicensure}{\textbf{Output:}}
\REQUIRE Unbounded External Archive {\bf UEA}, current population {\bf pop},, number of vectors to add {\bf nav}
\ENSURE  Updated population $\bf{pop}$
\STATE set count = 0
\WHILE{$count < \bf{nav}$}
\STATE generate new vector $\bf{\lambda^{rand}}$ using values sampled from a uniform distribution
\STATE set $\bf{ind^{rand}}$ =($\bf{x^{rand}},\bf{FV^{rand}}$) best solution for $\bf{\lambda^{rand}}$ in $\bf{UEA}$
\STATE add ($\bf{ind^{rand}}$,$\bf{\lambda^{rand}}$) to $\bf{pop}$
\STATE count = count+1
\ENDWHILE
\RETURN $\bf{pop}$
\end{algorithmic} 
\end{algorithm}

When adding new vectors, MOEA/D-AV has a choice of adding vectors using two methods: the first is based on AWA (Algorithm~\ref{algo::addVector}) and the second is based on uniform sampled values. This second method samples values from a uniform distribution and then generates new weight vectors (Algorithm~\ref{algo::addVector2}). This second method is used because determining the position of new vectors only with AWA leads to early stagnation of the search. 

The choice of which method to use is controlled by the $p$ probability value, which changes as the search progresses. This probability is calculated as $p = \frac{n\_fe}{n\_eval}$. Where $n\_fe$ is the current number of function evaluations, and $n\_eval$ is the total evaluation budget. This equation causes MOEA/D-AV to add weight vectors generated from values sampled from a uniform distribution at the beginning of the search. Then, this MOEA/D variant is more likely to create new weight vectors using the AWA-based method at later states.

\begin{algorithm}
\caption{Delete Vectors}
\label{algo::delete Vector}
\begin{algorithmic}[1]
\renewcommand{\algorithmicrequire}{\textbf{Input:}}
\renewcommand{\algorithmicensure}{\textbf{Output:}}
\REQUIRE Current population {\bf pop}, number of vectors to delete {\bf nav}
\ENSURE  Updated population $\bf{pop}$
\STATE set count = 0
\WHILE{$count < \bf{nav}$}
\STATE let $W_{notEdge}$ be $W$ without the edge vectors
\STATE randomly select {\bf argument} {\bf arg} from $W_{notEdge}$
\STATE delete $(\bf{ind^{arg}},\bf{\lambda^{arg}})$ from $\bf{pop}$
\STATE count = count+1
\ENDWHILE
\RETURN $\bf{pop}$
\end{algorithmic} 
\end{algorithm}

Besides adding weight vectors, MOEA/D-AV also deletes weight vectors, with the goal of avoiding wasting computational resources when there are too many weight vectors in use. Currently, the weight vectors are selected to be deleted randomly, excluding those weight vectors associated with the axis of each objective. Algorithm~\ref{algo::delete Vector} describes the method in detail.
\section{Weight Vectors Experiment}

We investigate the relationship between adapting the number of weight vectors, represented by MOEA/D-AV, and increments of performance in practice. Here, we compare the UEA of MOEA/D-AV, MOEA/D-DE, and the MOEA/D-AWA (which adjust the values of the weight vectors, but not their numbers). The three methods are compared with different numbers of initial weight vectors to analyze how they interact given this different settings.

The algorithms were compared on the Hypervolume (HV)\footnote{For the HV calculation, we use the reference point set to (1 + 1/H,1 + 1/H) for two objective problems and (1 + 1/H, 1 + 1/H, 1 + 1/H) for three objective problems.}, IGD and Entropy metrics. The first two comparison methods are common in the MOP literature. This Entropy metric~\cite{entropy} is beneficial to evaluate the sparseness of the PF and it is uses to measure how well a method covers empty areas of the PF, the main motivation of this work. For the comparisons, we use the DTLZ benchmark set (3-objective, 10 dimensions)~\cite{dtlz}, and the ZDT set (2-objective, 30 dimensions)~\cite{ZDT}. For a fair comparison, the algorithms are evaluated based on their Unbounded External Archive (UEA)~\cite{benchmarking}.


The general MOEA/D-DE parameters were are used here as they were introduced in~\cite{moead_de}. On the other hand, the Generation gap and user-controlled fraction are the same as from the CR paper~\cite{cr}. Also, specific parameters of the MOEA/D-AWA can be found at~\cite{moead_awa}. Finally, our MOEA/D-AV adds one new parameter, the vector adaption ratio. The number of weight vectors for the 3-objective DTLZ benchmark set was selected to be \{10, 21, 45, 105, 190, 496 and 990\}. The number of weight vectors for the 2-objective ZDT set was set to \{10, 20, 50, 100, 200, 500 and 1000\}~\footnote{We initialize the weight vectors using the SLD method, causing the number to slightly change between MOPs with 2 and 3 objectives.}. We set the number of evaluations to 75000 and the number of trials to 21. The difference in performance for each experimental condition, across all benchmark sets, was evaluated using a two-sided Wilcoxon signed-rank test paired by benchmark, with $\alpha = 0.05$.

\section{Results}

\begin{table*}[htbp]
\centering
    \caption{Mean and standard deviation, in parenthesis, for all algorithms.} 
    \label{tab::Best_Scenario}
    \scalebox{1}{
    \begin{tabular}{c|ccc|ccc}
    \hline \multicolumn{4}{c|}{\textbf{Best scenario}}& \multicolumn{3}{c}{\textbf{Worse scenario}} \\ \hline 
    $\bf{HV}$ & MOEA/D  & AWA & \textbf{MOEA/D-AV}& MOEA/D  & AWA & \textbf{MOEA/D-AV} \\ \hline 
    DTLZ1 & 0.97 (0.02) & \cellcolor[gray]{0.85} 0.98 (0.00) & 0.95 (0.06) & 0.21 (0.27) & 0.25 (0.33) & \cellcolor[gray]{0.85}0.51 (0.42) \\ 
    DTLZ2 & \cellcolor[gray]{0.85} 0.47 (0.00) & \cellcolor[gray]{0.85} 0.47 (0.00) & \cellcolor[gray]{0.85} 0.47 (0.00) & 0.45 (0.00) & 0.45 (0.00) & \cellcolor[gray]{0.85}0.47 (0.00) \\ 
    DTLZ3 & \cellcolor[gray]{0.85} 0.45 (0.01) & \cellcolor[gray]{0.85} 0.45 (0.01) & 0.44 (0.02) & 0.01 (0.03) & 0.02 (0.04) & \cellcolor[gray]{0.85}0.15 (0.16) \\ 
    DTLZ4 & \cellcolor[gray]{0.85} 0.46 (0.00)
 & \cellcolor[gray]{0.85} 0.46 (0.00) & \cellcolor[gray]{0.85} 0.46 (0.00) & 0.33 (0.12) & 0.33 (0.15) & \cellcolor[gray]{0.85}0.44 (0.05) \\  
    DTLZ5 & \cellcolor[gray]{0.85} 0.46 (0.00)& \cellcolor[gray]{0.85} 0.22 (0.00) & \cellcolor[gray]{0.85} 0.22 (0.00) & \cellcolor[gray]{0.85}0.22 (0.00) & \cellcolor[gray]{0.85}0.22 (0.00) & \cellcolor[gray]{0.85}0.22 (0.00) \\ 
    DTLZ6 & \cellcolor[gray]{0.85} 0.22 (0.00) & \cellcolor[gray]{0.85} 0.22 (0.00) & \cellcolor[gray]{0.85} 0.22 (0.00) & 0.19 (0.08) & \cellcolor[gray]{0.85}0.22 (0.00) & \cellcolor[gray]{0.85}0.22 (0.00) \\ 
    DTLZ7 & \cellcolor[gray]{0.85} 0.24 (0.00) & \cellcolor[gray]{0.85} 0.24 (0.00)  & \cellcolor[gray]{0.85} 0.24 (0.00)& 0.13 (0.03) & 0.21 (0.03) & \cellcolor[gray]{0.85}0.22 (0.04) \\ 

ZDT1 & 0.66 (0.00) & 0.66 (0.00) & \cellcolor[gray]{0.85}0.67 (0.00) & 0.03 (0.03) & 0.03 (0.02) & \cellcolor[gray]{0.85}0.12 (0.05) \\ 
ZDT2 & \cellcolor[gray]{0.85} 0.33 (0.00) & \cellcolor[gray]{0.85}0.33 (0.00)& \cellcolor[gray]{0.85}0.33 (0.00)& \cellcolor[gray]{0.85}0.00 (0.00) & \cellcolor[gray]{0.85}0.00 (0.00) & \cellcolor[gray]{0.85}0.00 (0.00) \\  
ZDT3 & \cellcolor[gray]{0.85}1.04 (0.00) & \cellcolor[gray]{0.85}1.04 (0.00) & \cellcolor[gray]{0.85}1.04 (0.00) & 0.20 (0.05) & 0.23 (0.04) & \cellcolor[gray]{0.85}0.33 (0.05) \\  
ZDT4 & \cellcolor[gray]{0.85}0.66 (0.00) & \cellcolor[gray]{0.85}0.66 (0.00) & \cellcolor[gray]{0.85}0.66 (0.00) & \cellcolor[gray]{0.85}0.00 (0.00) & \cellcolor[gray]{0.85}0.00 (0.00) & \cellcolor[gray]{0.85}0.00 (0.00) \\ 
ZDT6 & \cellcolor[gray]{0.85}0.33 (0.00) & \cellcolor[gray]{0.85}0.33 (0.00) & \cellcolor[gray]{0.85}0.33 (0.00)  & \cellcolor[gray]{0.85}0.00 (0.00) & \cellcolor[gray]{0.85}0.00 (0.00) & \cellcolor[gray]{0.85}0.00 (0.00) \\ 
\hline Stats & $=$ & $=$ & & $+$ & $+$ &  \\  

\hline $\bf{IGD}$ & MOEA/D  & AWA & \textbf{MOEA/D-AV}& MOEA/D  & AWA & \textbf{MOEA/D-AV} \\  \hline 

DTLZ1 & 0.58 (0.13) & 0.58 (0.17) & \cellcolor[gray]{0.85} 0.57 (0.19) & 1.33 (0.92) & 4.38 (6.26) & \cellcolor[gray]{0.85}0.96 (1.05) \\  
DTLZ2 & 0.01 (0.00) & 0.01 (0.00) & \cellcolor[gray]{0.85} 0.00 (0.00) & \cellcolor[gray]{0.85}0.01 (0.00) & \cellcolor[gray]{0.85}0.01 (0.00) & \cellcolor[gray]{0.85}0.01 (0.00) \\ 
DTLZ3 & \cellcolor[gray]{0.85}0.01 (0.01) & \cellcolor[gray]{0.85}0.01 (0.00) & 0.02 (0.02) & 4.98 (5.45) & 6.96 (7.86) & \cellcolor[gray]{0.85}2.14 (2.85) \\ 
DTLZ4 & \cellcolor[gray]{0.85}0.01 (0.00) & \cellcolor[gray]{0.85}0.01 (0.00) & \cellcolor[gray]{0.85}0.01 (0.00) & 0.33 (0.31) & 0.29 (0.34) & \cellcolor[gray]{0.85}0.06 (0.16) \\ 
DTLZ5 & \cellcolor[gray]{0.85}0.00 (0.00) & \cellcolor[gray]{0.85}0.00 (0.00) & \cellcolor[gray]{0.85}0.00 (0.00)& \cellcolor[gray]{0.85}0.00 (0.00) & \cellcolor[gray]{0.85}0.00 (0.00) & \cellcolor[gray]{0.85}0.00 (0.00) \\
DTLZ6 & \cellcolor[gray]{0.85}0.00 (0.00) & \cellcolor[gray]{0.85}0.00 (0.00) & \cellcolor[gray]{0.85}0.00 (0.00) & 0.14 (0.34) & \cellcolor[gray]{0.85}0.00 (0.00) & \cellcolor[gray]{0.85}0.00 (0.00) \\ 
DTLZ7 & \cellcolor[gray]{0.85}0.01 (0.00) & \cellcolor[gray]{0.85}0.01 (0.00) & \cellcolor[gray]{0.85}0.01 (0.00) & 0.39 (0.17) & \cellcolor[gray]{0.85}0.13 (0.17) & 0.15 (0.27) \\  
ZDT1 & \cellcolor[gray]{0.85}0.00 (0.00) & \cellcolor[gray]{0.85}0.00 (0.00) & \cellcolor[gray]{0.85}0.00 (0.00) & 0.73 (0.13) & 0.69 (0.07) & \cellcolor[gray]{0.85}0.50 (0.09) \\  
ZDT2 & \cellcolor[gray]{0.85}0.00 (0.00) & \cellcolor[gray]{0.85}0.00 (0.00) & \cellcolor[gray]{0.85}0.00 (0.00) & 1.29 (0.14) & 1.18 (0.19) & \cellcolor[gray]{0.85}0.91 (0.18) \\ 
ZDT3 & \cellcolor[gray]{0.85}0.00 (0.00) & \cellcolor[gray]{0.85}0.00 (0.00) & \cellcolor[gray]{0.85}0.00 (0.00) & 0.61 (0.08) & 0.57 (0.06) & \cellcolor[gray]{0.85}0.46 (0.05) \\  
ZDT4 & \cellcolor[gray]{0.85}0.00 (0.00) & \cellcolor[gray]{0.85}0.00 (0.00) & \cellcolor[gray]{0.85}0.00 (0.00) & 21.1 (4.34) & 20.5 (4.33) & \cellcolor[gray]{0.85}17.5 (3.16) \\ 
ZDT6 & \cellcolor[gray]{0.85}0.00 (0.00) & \cellcolor[gray]{0.85}0.00 (0.00) & \cellcolor[gray]{0.85}0.00 (0.00) & 5.64 (0.27) & 5.55 (0.36) & \cellcolor[gray]{0.85}5.02 (0.51) \\ 
\hline Stats & $=$ & $=$ &  & $+$ & $+$ &  \\ 
\hline $\bf{Entropy}$ & MOEA/D & AWA & \textbf{MOEA/D-AV}& MOEA/D  & AWA & \textbf{MOEA/D-AV} \\ \hline  

DTLZ1 & 11.2 (0.88) & \cellcolor[gray]{0.85} 11.4 (0.73) & 10.3 (1.15)& 3.75 (0.45) & 3.74 (0.59) & \cellcolor[gray]{0.85} 7.28 (2.04) \\ 
DTLZ2 &  12.2 (0.02)& \cellcolor[gray]{0.85} 12.3 (0.03) & 12.2 (0.04)& 9.19 (0.16) & 8.62 (0.45) & \cellcolor[gray]{0.85} 11.1 (0.31) \\ 
DTLZ3 & 11.0 (0.72)& \cellcolor[gray]{0.85} 11.2 (0.58) & 10.1 (1.28) & 2.67 (0.48) & 1.83 (0.88) & \cellcolor[gray]{0.85} 5.08 (2.58) \\ 
DTLZ4 & 11.1 (0.16) & \cellcolor[gray]{0.85} 11.1 (0.17) & 10.9 (0.11)& 6.00 (1.49) & 5.47 (2.35) & \cellcolor[gray]{0.85} 8.32 (0.84) \\ 
DTLZ5 & \cellcolor[gray]{0.85} 6.98 (0.03) & \cellcolor[gray]{0.85} 6.98 (0.02) & 6.92 (0.02)& 6.45 (0.05) & 5.71 (0.24) & \cellcolor[gray]{0.85}6.66 (0.03) \\ 
DTLZ6 & 6.58 (0.03) & 6.59 (0.00) & \cellcolor[gray]{0.85} 6.60 (0.02)& \cellcolor[gray]{0.85} 6.44 (0.31) & 5.29 (0.24) & 6.17 (0.08) \\ 
DTLZ7 & 10.1 (0.08) & 10.1 (0.14) &  \cellcolor[gray]{0.85}10.2  (0.25)& 8.79 (0.91) & 8.04 (0.70) & \cellcolor[gray]{0.85} 9.22 (0.44) \\  
ZDT1 & \cellcolor[gray]{0.85} 7.26 (0.03) & 7.25 (0.03) & 7.12 (0.22) & 6.02 (0.27) & 5.91 (0.28) & \cellcolor[gray]{0.85}6.19 (0.20) \\ 
ZDT2 & 7.24 (0.04) & \cellcolor[gray]{0.85} 7.25 (0.09) & 7.18 (0.17)& 2.04 (1.50) & 2.45 (1.47) & \cellcolor[gray]{0.85} 3.38 (1.41) \\ 
ZDT3 & \cellcolor[gray]{0.85} 6.66 (0.06) & 6.65 (0.05) & 6.56 (0.50)& 5.96 (1.60) & 6.02 (0.18) & \cellcolor[gray]{0.85} 6.11 (0.11) \\ 
ZDT4 & 7.20 (0.06) & 7.26 (0.02) & \cellcolor[gray]{0.85} 7.27 (0.08) & 2.67 (0.54) & 6.78 (0.44) & \cellcolor[gray]{0.85} 6.98 (0.25) \\ 
ZDT6 & 6.62 (0.24) & \cellcolor[gray]{0.85} 6.71 (0.29) & 4.30 (0.37) & 3.93 (0.47) & \cellcolor[gray]{0.85} 4.09 (0.61) & 2.64 (1.07) \\ 
\hline Stats & $-$ & $-$ & $+$ & $+$ &  \\ \hline
    
    \end{tabular}
    }
\end{table*}

\begin{table}[htbp]

\centering
    \caption{Paring of initial number of weight vectors and MOEA/D variant used for comparison of the ``Best Scenario" (Table~\ref{tab::Best_Scenario}) and ``Worst Scenario" (Table~\ref{tab::Worst_Scenario}). These were selected using the best and worst mean HV values, respectively.}
    \label{tab::number of vector}
    \scalebox{1}{
    \begin{tabular}{l|c|c|c|c|c|c}
    \multicolumn{7}{c}{\bf{Best Scenario --- Worst scenario}} \\ \hline& MOEA/D  & AWA & \textbf{MOEA/D-AV} & MOEA/D  & AWA & \textbf{MOEA/D-AV} \\ \hline
    DTLZ1 & 190 & 105 & 190 & 990 & 990 & 990 \\
    DTLZ2 & 190 & 190 & 10 & 10 & 10 & 990 \\
    DTLZ3 & 190 & 105 & 190 & 990 & 990 & 990 \\
    DTLZ4 & 496 & 496 & 990 & 10 & 10 & 10 \\
    DTLZ5 & 190 & 105 & 21 & 21 & 990 & 990 \\
    DTLZ6 & 496 & 496 & 190 & 10 & 10 & 990 \\
    DTLZ7 & 496 & 190 & 10 & 10 & 10 & 21 \\ 
    ZDT1 & 50 & 10 & 100 & 1000 & 1000 & 1000 \\
    ZDT2 & 10 & 10 & 20 & 1000 & 1000 & 1000 \\
    ZDT3 & 50 & 50 & 10 & 1000 & 1000 & 1000 \\
    ZDT4 & 100 & 100 & 20 & 500 & 500 & 1000 \\
    ZDT6 & 50 & 50 & 100 & 500 & 1000 & 1000 \\
    \end{tabular}
    }
\end{table}

This section compares MOEA/D-AV against the traditional MOEA/D-DE and MOEA/D-AWA (AWA for simplicity) with different numbers of weight vectors . We recall that the algorithms are evaluated based on their Unbounded External Archive (UEA) and not their final population~\cite{benchmarking}, for fair comparisons. The results of the statistical tests (Wilcoxon signed-rank test paired by benchmark, with $\alpha = 0.05$) are shown on Table~\ref{tab::Best_Scenario}. In this work, we use the symbols ``=", ``+" and ``-" to indicate the results of the statistical test. The symbol ``=" indicates no statistically significant difference between the methods, while ``+" is used to indicate a significant difference in favour of MOEA/D-AV and ``-" indicates difference against MOEA/D-AV. 

\begin{figure*}[htbp]
\subfloat[MOEA/D-AV method finds better HV values in most initial settings. We highlight the results of MOEA/D-AV with only 10 initial vectors.]{
\includegraphics[width=0.45\textwidth]
{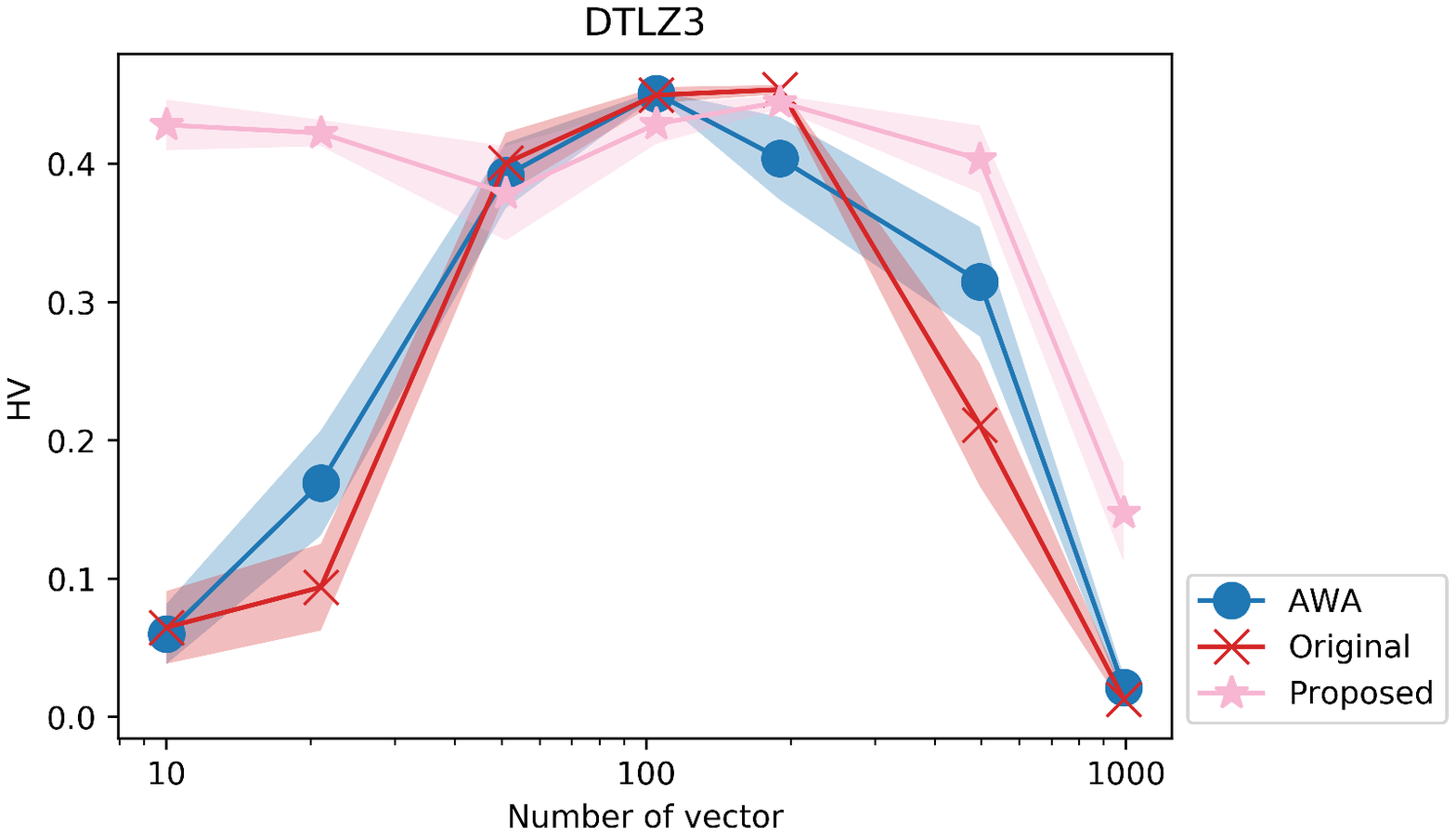}}
\quad
\subfloat[ MOEA/D-AV has the best performance at lower number of initial weight vectors and has competitive results for all settings. ]{
\includegraphics[width=0.45\textwidth]
{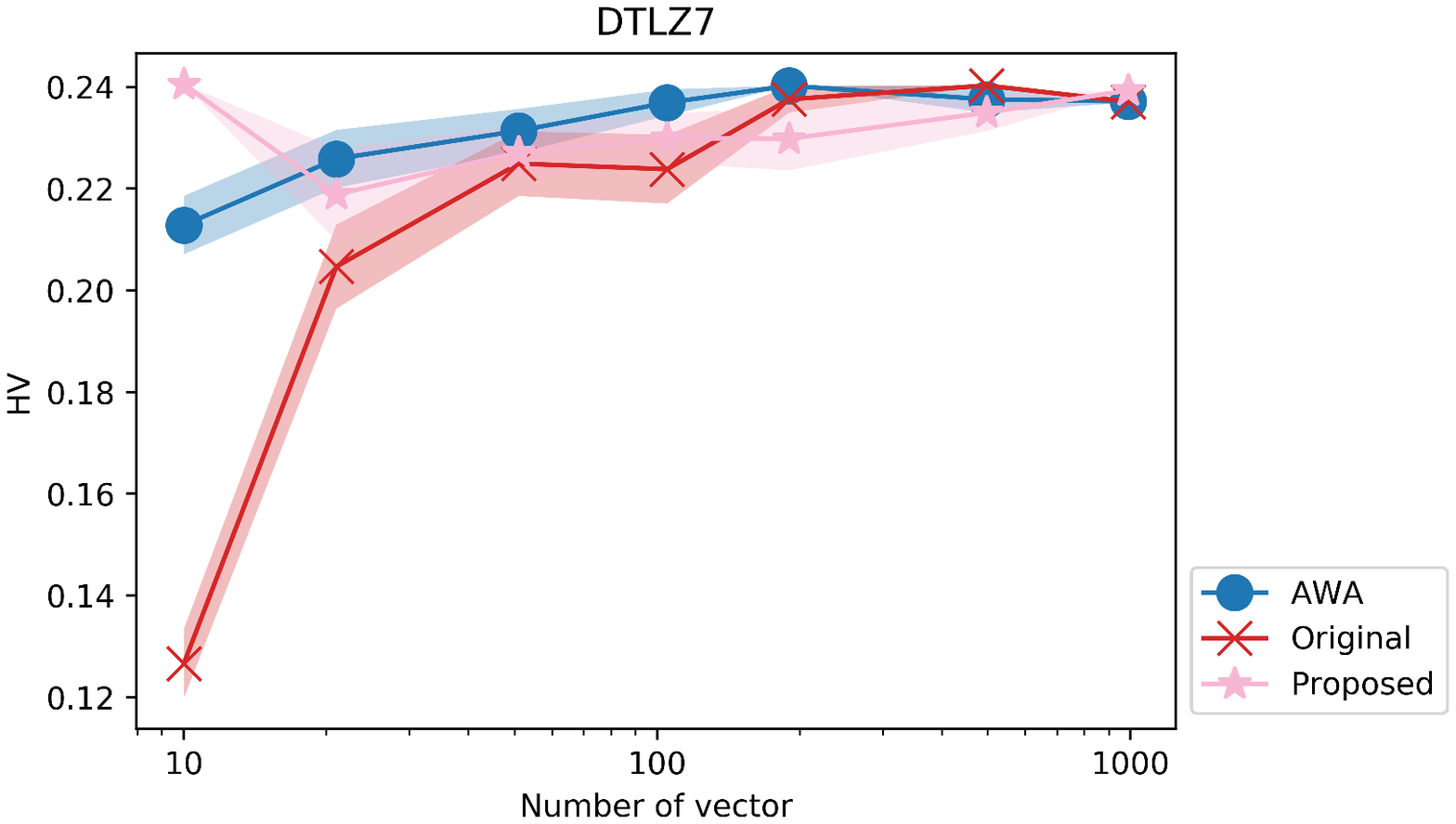}}
\caption{Mean HV value against initial number of weight vectors, for DTLZ3 on the left and DTLZ7 on the right. Shaded areas indicate standard deviations.}
\label{fig::robust}
\end{figure*}

Table~\ref{tab::Best_Scenario}, left side, shows the mean and standard deviation of the \textit{best setting for each algorithm in terms of the number of weight vectors}, based on hypervolume values. The best method for each MOP is highlighted in bold. Looking at this Table, we can see that the best results are similar in terms of HV and IGD. This result suggests that there is no apparent difference between these methods. In terms of Entropy values, the proposed method performs a little worse than the other methods, especially for DTLZ1-4. On the other hand, Table~\ref{tab::Best_Scenario}, right side, shows the mean and standard deviation of the \textit{worst setting for each algorithm in terms of the number of weight vectors}. The results are shown by this side of the Table~\ref{tab::Best_Scenario} indicating that the MOEA/D-AV performs better than the other two MOEA/D variants in all metrics. It is in our understanding that the reason for this is that our method can compensate for initial bad choices of the number of vectors and achieve competitive results. Finally, Table~\ref{tab::number of vector} shows the number of vectors for best and worst settings scenarios in terms of HV.


It is interesting to note that using extremes values for the number of weight vectors, such as 10 or 990, lead to bad HV performance, as can be seen in Table~\ref{tab::number of vector}. In the case of a higher number of weight vectors, a possible cause for this low performance is due to the number of vectors being too large for the algorithms to efficiently progress with the search progress. In the case of the lower number of weight vectors, the reason for its bad performance may be that such configuration provides little information about the search progress.

\begin{figure*}[htbp]
\subfloat[MOEA/D-DE with 21 vectors]{
\includegraphics[width=0.30\textwidth]
{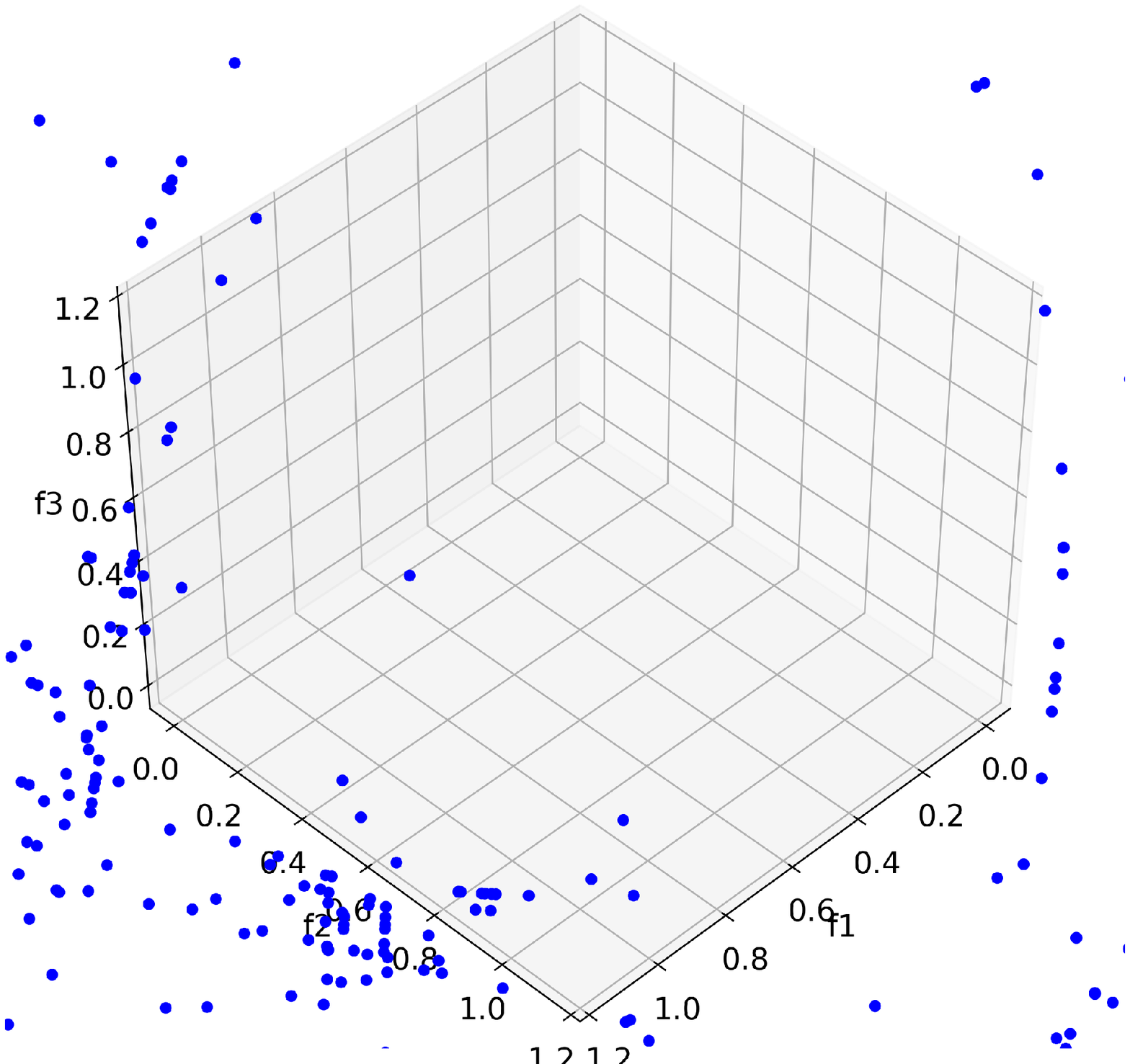}}\quad
\subfloat[MOEA/D-AWA with 21 vectors]{
\includegraphics[width=0.30\textwidth]
{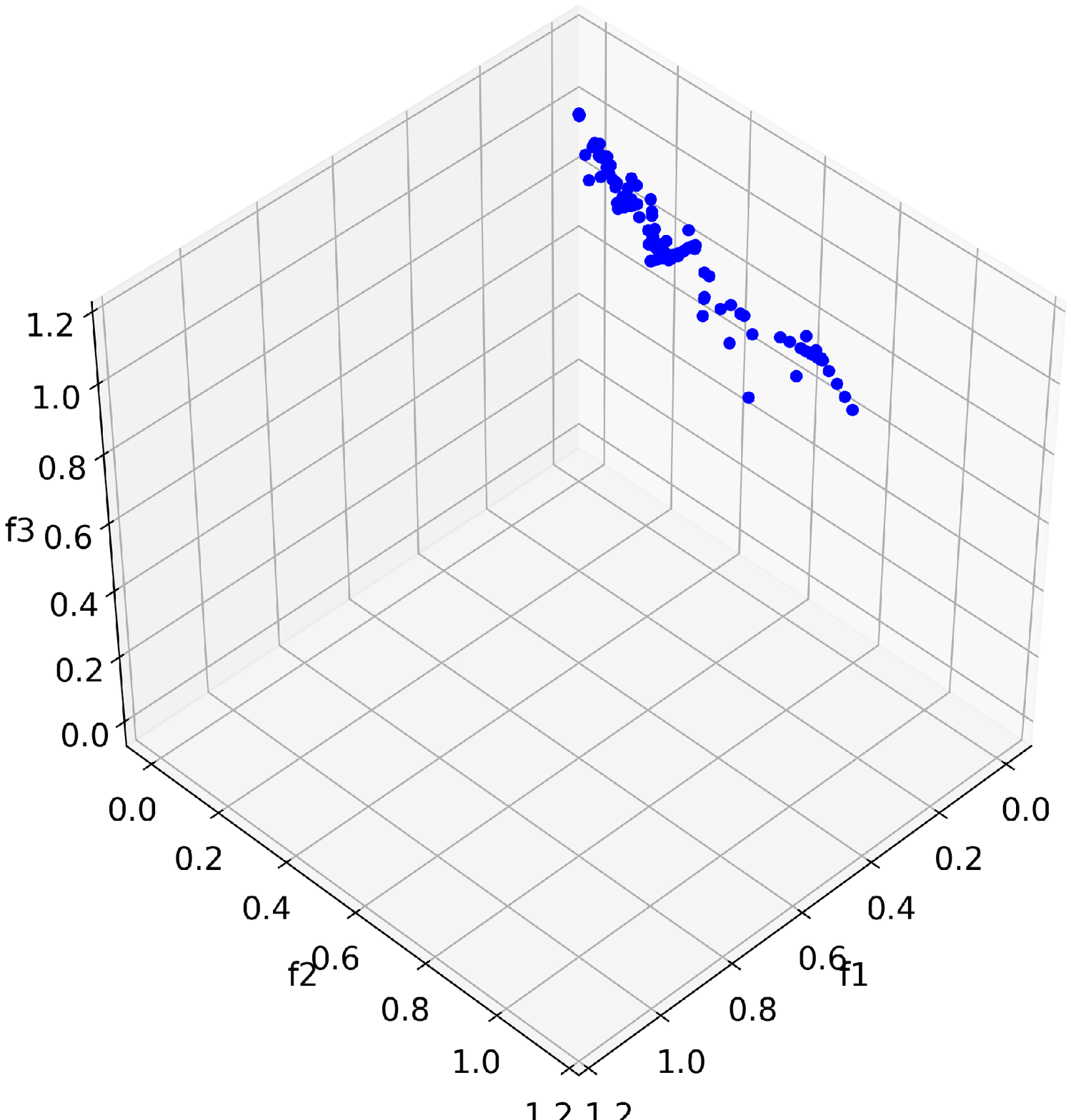}}\quad
\subfloat[MOEA/D-AV with 21 vectors]{
\includegraphics[width=0.30\textwidth]
{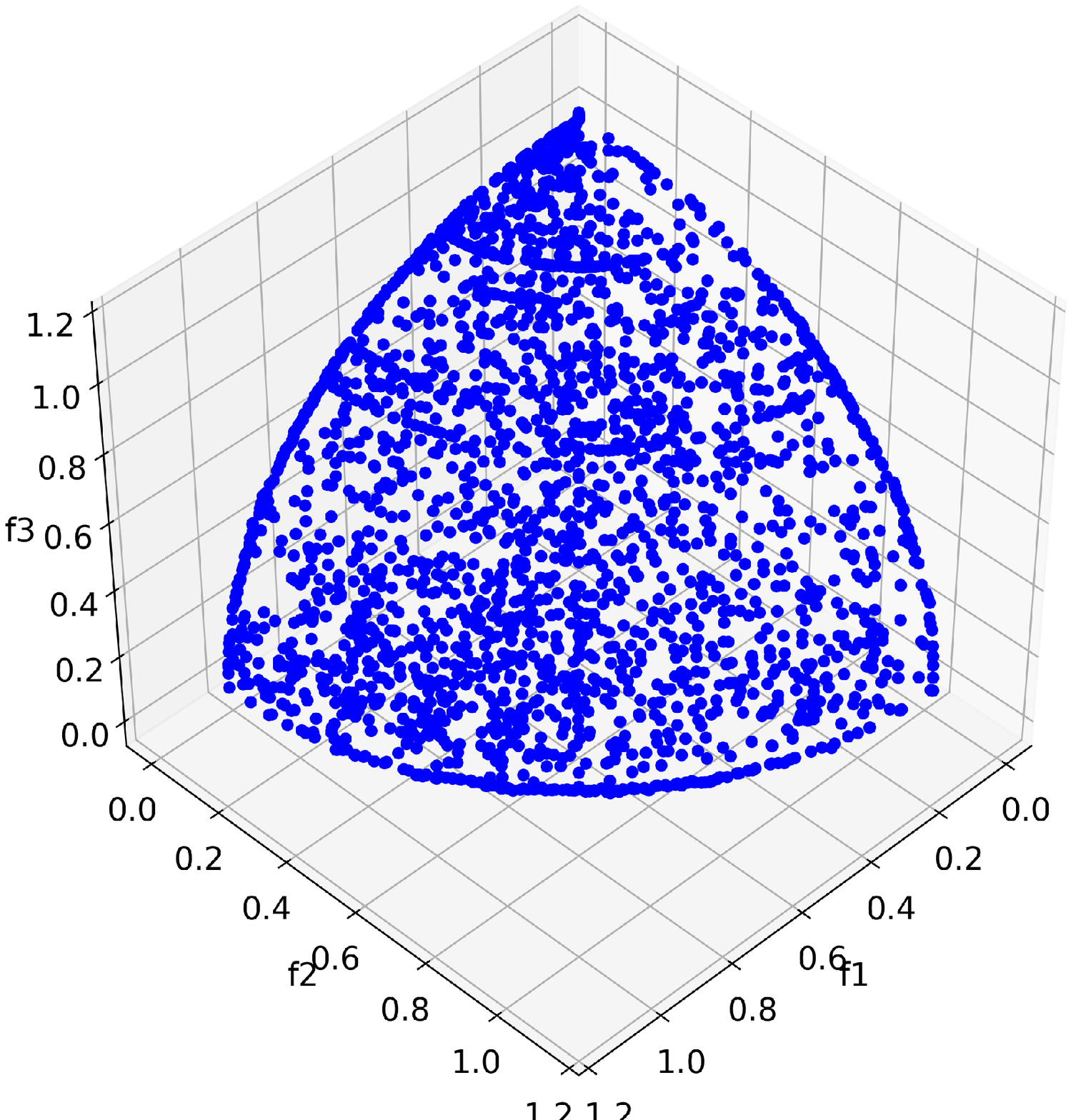}}
\caption{UEA of 3 methods starting from 21 vectors in DTLZ3. MOEA/D-DE and MOEA/D-AWA have low coverage of the PF, while MOEA/D-AV is able to cover well most regions of the objective space.}
\label{fig::UEA of DTLZ3_pop21}
\end{figure*}

\begin{figure*}[htbp]
    \centering
    \subfloat[MOEA/D-DE with 496 vectors]{
    \includegraphics[width=0.30\textwidth]
    {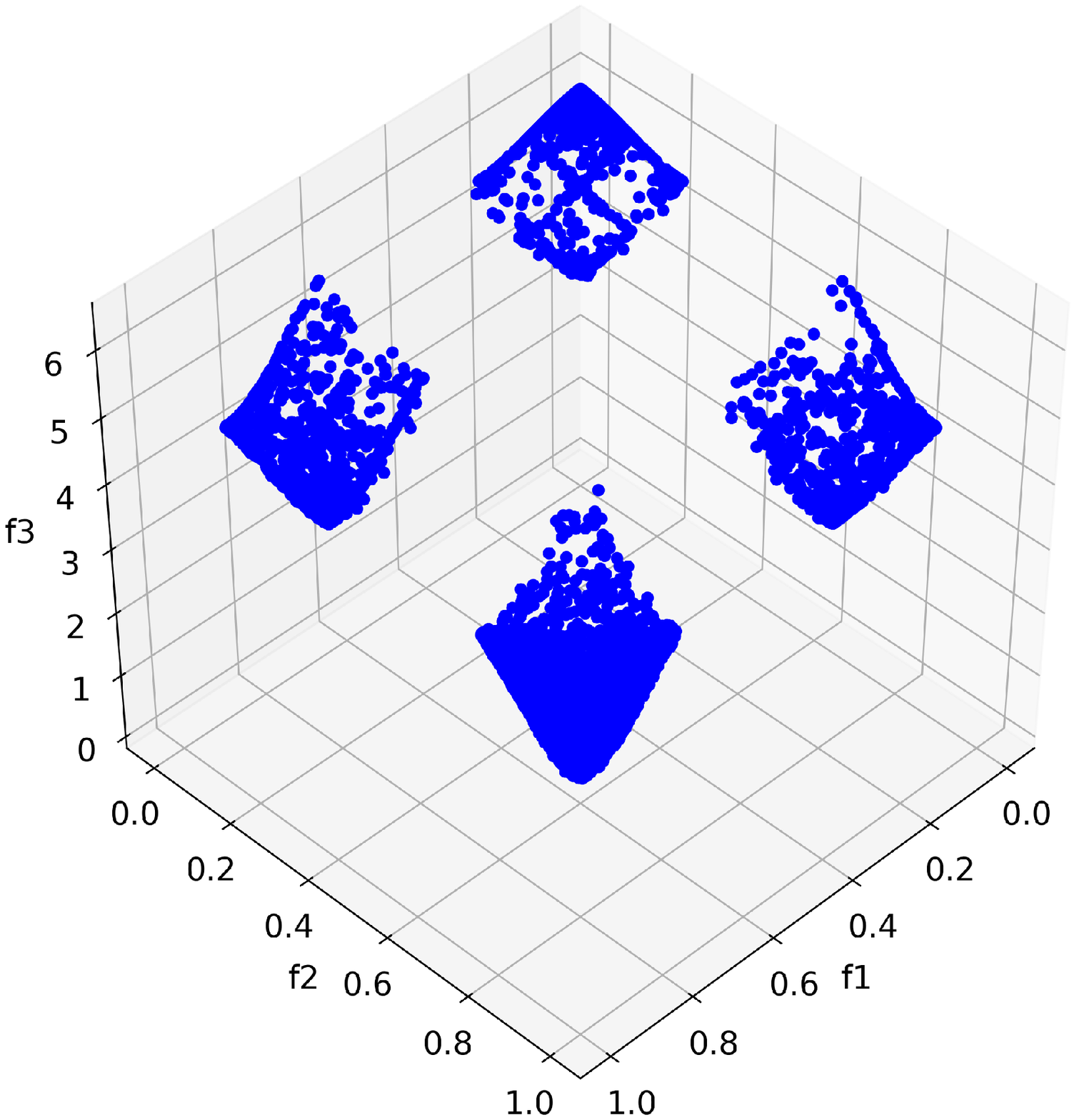}}\quad
    \subfloat[MOEA/D-AWA with 496 vectors]{
    \includegraphics[width=0.30\textwidth]
    {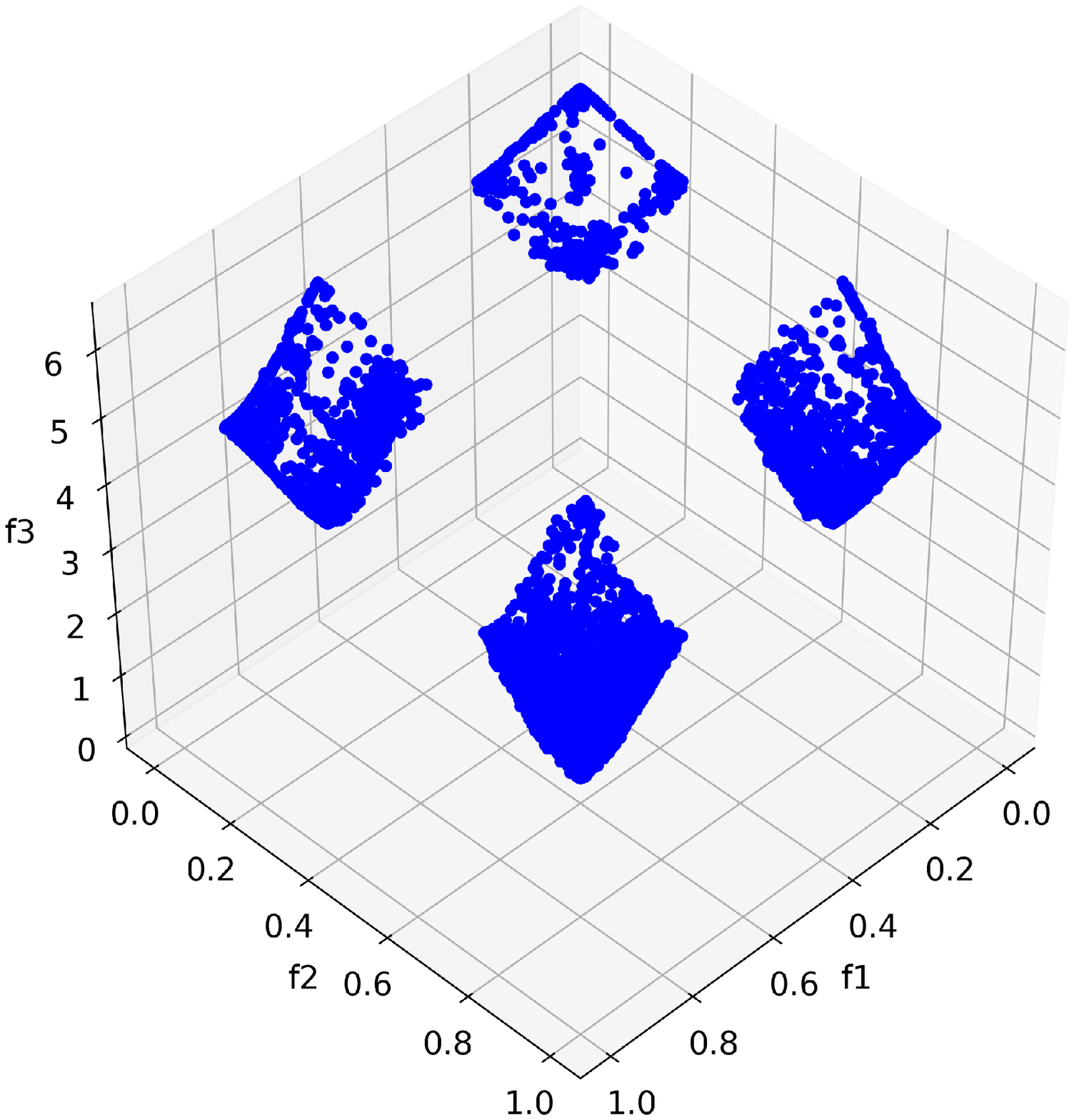}}\quad
    \subfloat[MOEA/D-AV with 496 vectors]{
    \includegraphics[width=0.30\textwidth]
    {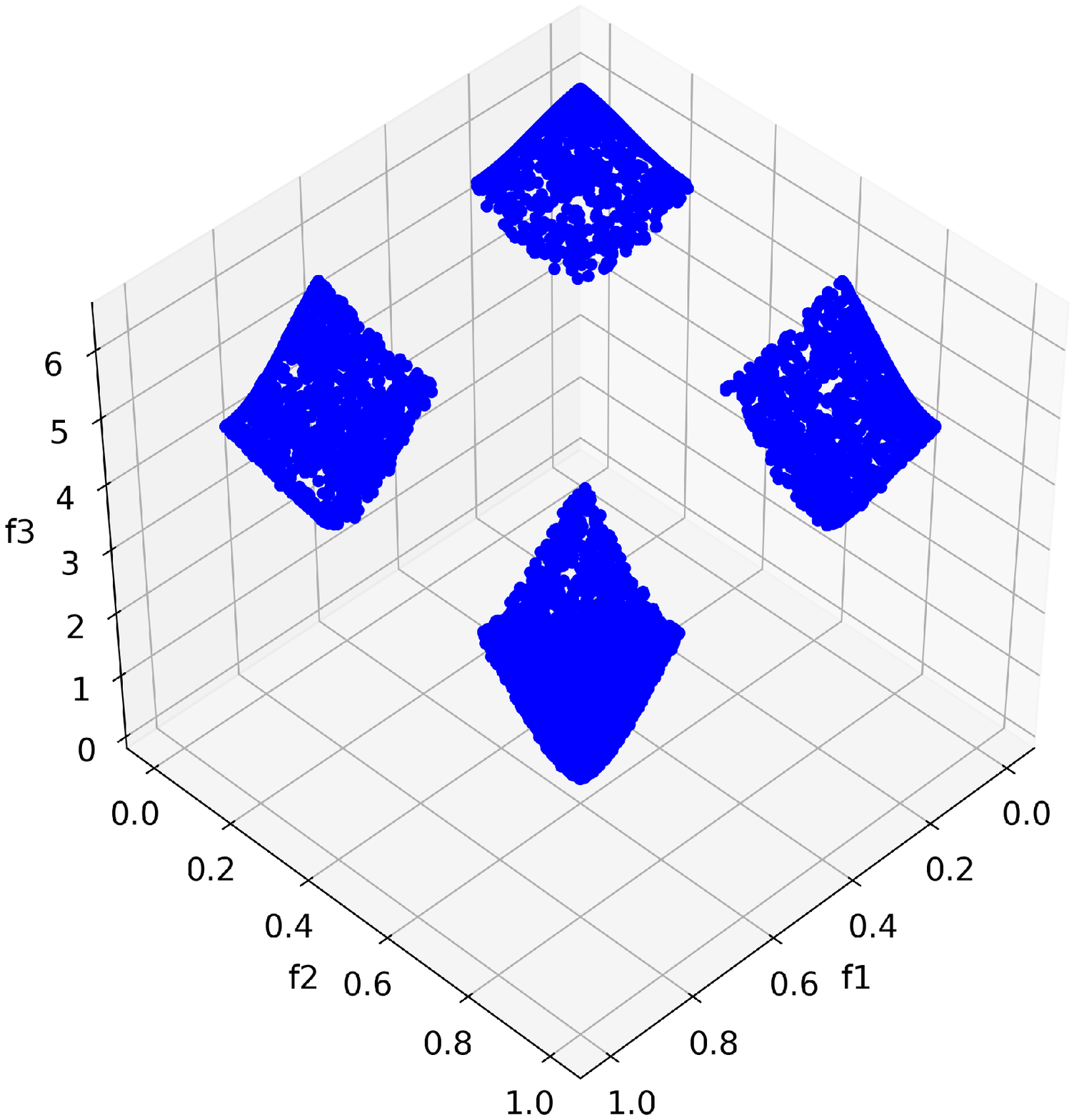}}
    \caption{UEA of the 3 methods starting from 105 vectors in DTLZ7. Although there is little difference in the HV values for each of the methods, we can clearly see that MOEA/D-AV can fill empty regions of the objective space evenly.}
    \label{fig::UEA of DTLZ_pop105}
\end{figure*}

\begin{figure*}[htbp]
\subfloat[Change in the number of vectors of MOEA/D-AV in DTLZ1. Best setting (blue, continuous line) starts with 190 vectors and the worst setting (brown, dashed line) starts with 990 vectors.]{
\includegraphics[width=0.45\textwidth]
{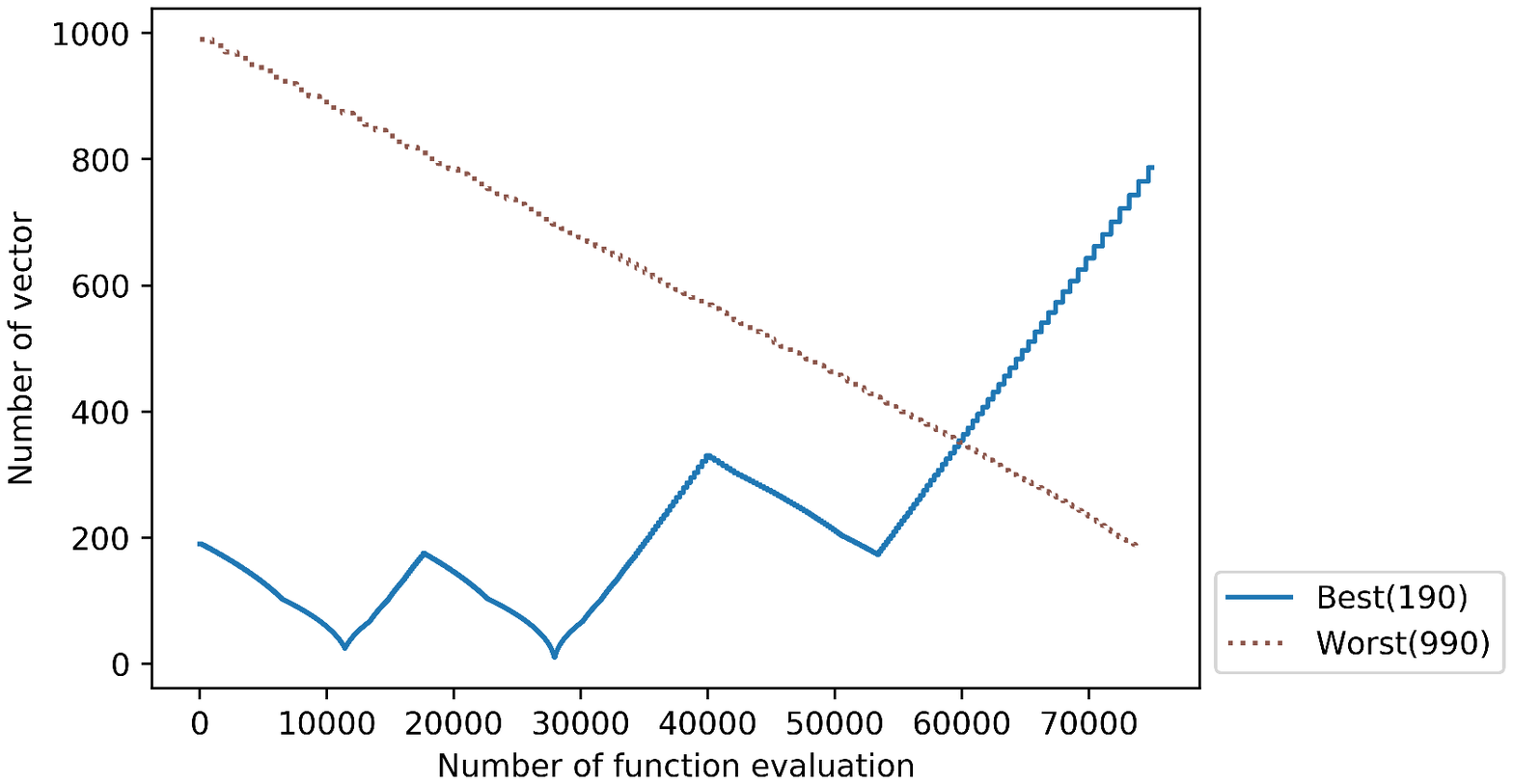}}
\quad
\subfloat[Change in the number of vectors of MOEA/D-AV in ZDT2. Best setting (blue, continuous line) starts with 10 vectors and the worst setting (brown, dashed line) starts with 1000 vectors.]{
\includegraphics[width=0.45\textwidth]
{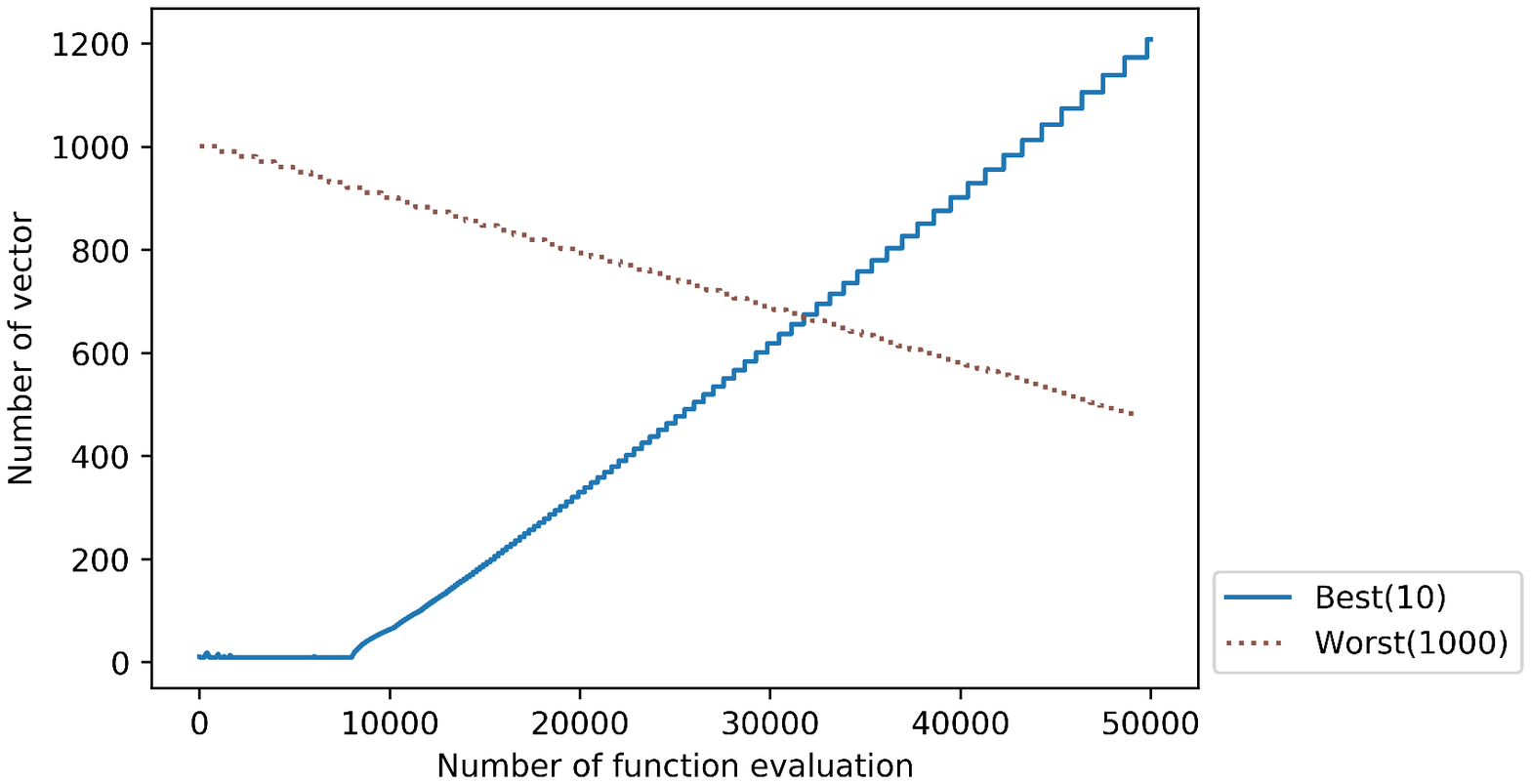}}
\caption{MOEA/D-AV works better with low number of weight vectors, while higher values cause MOEA/D-AV to reduce the number of vectors.}
\label{fig::vectors}
\end{figure*}

Figure~\ref{fig::robust} shows the change in the HV of the solutions of the final UEA for each method, on the  DTLZ3 with continuous PF (a - left side) and DTLZ7 with discontinuous PF, (b - right side). On DTLZ3, MOEA/D-AV achieves higher or competitive results independently of the initial number of vectors and the performance of both MOEA/D-DE and MOEA/D-AWA deteriorates significantly when the number of weight vectors is not set correctly. 

Figure~\ref{fig::UEA of DTLZ3_pop21} depicts the PF approximated for each method on DTLZ3. When the number of weight vectors is small, only MOEA/D-AV can provide a suitable approximation to the PF. We believe that reason why the other algorithms behave badly in this case is because the distance of the weight vectors provides little useful information about the search progress. This result supports the need to add vectors randomly to avoid early stagnation of the search, a feature only present in MOEA/D-AV.

Coming back to Figure~\ref{fig::robust}, we discuss the results of all algorithms in DTLZ7. MOEA/-AV is the only algorithm to achieve good results independently of the number of vectors. Although the HV performance of the methods is similar, Figure~\ref{fig::UEA of DTLZ_pop105} shows that their ability to cover the PF differs. MOEA/D-AV method provides a wider coverage of the optimal PF and this is related to the ability of this algorithm to add vectors to empty areas of the objective space and to remove weight vectors in areas of the PF already covered.



Figure~\ref{fig::vectors} shows the change in the number of weight vectors for the best and worst setting scenario in DTLZ1 and ZDT2, respectively. At Figure~\ref{fig::vectors} (a - left side) the blue, continuous line illustrates that MOEA/D-AV to keep reducing and increasing the number of weight vectors until around 50000 evaluations. Then, the algorithm seems to keep increasing the weight vectors. On the other hand, Figure~\ref{fig::vectors} (b - right side) shows that MOEA/D-AV improves its coverage of the PF a little earlier than before, at 10000 evaluations. In both cases the worst case keeps reducing the number of weight vectors, confirming that starting with high number of vectors deteriorates the performance of any MOEA/D. 
\section{Conclusion}

Here we study the effect of adapting the number of weight vectors by removing unnecessary vectors and adding new vectors in empty areas of the PF. We proposed an the MOEA/D-AV that adaptively changes the number of weight vectors. This algorithm detects when the number of vectors must be changed and generates new vectors depending on the optimisation stage.

This study has shown that MOEA/D-AV has competitive performance, independently of the number of initial vectors. This result confirms that MOEA/D-AV finds suitable PF even when the number of initial vectors is not appropriate. One of the more significant findings is that using this method allows the use of MOEA/Ds without any fine-tuning process to choose the best set of the initial number of weight vectors. Thus, we understand that the dynamic adaptation of the number of weight vectors is a finding of interest for the whole EMO community. Future works include applying MOEA/D-AV in real-world MOPs, especially with non-regular and inverted PFs. Moreover, we want to compare MOEA/D-AV and MOEA/D with Resource Allocation methods that activate and deactivate vectors during the search~\cite{lavinas2020moea}.

\bibliographystyle{splncs04}
\bibliography{paper} 

\end{document}